\documentclass[journal]{IEEEtran}

\usepackage[draft]{hyperref}

\usepackage{graphicx}
\usepackage[font=small]{caption}
\usepackage{multirow}
\usepackage{hyperref}

\ifCLASSINFOpdf
\else
\fi
\begin{document}
\title{Distance-Based Anomaly Detection for Industrial Surfaces Using Triplet Networks}

% \author{Tareq Tayeh,~\IEEEmembership{Member,~IEEE,}
%         Sulaiman Aburakhia,~\IEEEmembership{Member,~IEEE,}
%         Abdallah Shami,~\IEEEmembership{Member,~IEEE}% <-this % stops a space
\author{Tareq Tayeh\IEEEauthorrefmark{1},
        Sulaiman Aburakhia\IEEEauthorrefmark{1},
        Ryan Myers\IEEEauthorrefmark{2},
        and Abdallah Shami\IEEEauthorrefmark{1}\\
        ECE Department, Western University, London, Canada \IEEEauthorrefmark{1}, National Research Council Canada, London, Canada \IEEEauthorrefmark{2}\\
\{ttayeh, saburakh, abdallah.shami\}@uwo.ca,\\
ryan.myers@nrc-cnrc.gc.ca}

\IEEEoverridecommandlockouts
\IEEEpubid{\makebox[\columnwidth]{978-1-7281-8416-6/20/\$31.00~\copyright2020 IEEE \hfill} \hspace{\columnsep}\makebox[\columnwidth]{ }}


\maketitle

\begin{abstract}
Surface anomaly detection plays an important quality control role in many manufacturing industries to reduce scrap production. Machine-based visual inspections have been utilized in recent years to conduct this task instead of human experts. In particular, deep learning Convolutional Neural Networks (CNNs) have been at the forefront of these image processing-based solutions due to their predictive accuracy and efficiency. Training a CNN on a classification objective requires a sufficiently large amount of defective data, which is often not available. In this paper, we address that challenge by training the CNN on surface texture patches with a distance-based anomaly detection objective instead. A deep residual-based triplet network model is utilized, and defective training samples are synthesized exclusively from non-defective samples via random erasing techniques to directly learn a similarity metric between the same-class samples and out-of-class samples. Evaluation results demonstrate the approach's strength in detecting different types of anomalies, such as bent, broken, or cracked surfaces, for known surfaces that are part of the training data and unseen novel surfaces.
\end{abstract}

% Evaluation results are very promising for known surfaces that are part of training and novel surfaces that are not part of training, where both contain many different defect types such as bent, broken, or contaminated surfaces. It demonstrates the approach's strength in detecting different types of anomalies on different industrial surfaces.

% The model achieves a defect type mean AUC score of 0.949 and a class mean AUC score of 0.960 when classifying the different defect types for known surfaces that are part of training and novel surfaces that are not part of training. The performance evaluations demonstrate the approach's strength in detecting different types of anomalies on different industrial surfaces.


% Novel Detection,

% \fi
\IEEEpeerreviewmaketitle

\section{Introduction}
Surface quality inspection is an essential part of the production process in all manufacturing industries around the globe. Scratched, bent, and other imperfect products may result in costly returns, imposing financial and operational issues \cite{CostlyReturns}. Quality control tasks rely largely on human experts who are trained on identifying anomalies, to ensure defective products are filtered out from the non-defective products that are ready to be used or shipped to consumers. However, as the production rates become faster and products become more complex, it becomes infeasible for humans to keep up with the throughput demand while trying to realize a near-perfect quality inspection. A 1\% productivity improvement across the industry can result in \$500 million in annual savings \cite{IntroData}. Furthermore, predicting anomalies on time can decrease over scheduled repairs by up to 12\%, maintenance costs by up to 30\%, and the number of breakdowns by up to 70\% \cite{IntroData}. As a result, machine-based visual inspections have been utilized in recent years to identify defective products, and in particular, Convolutional Neural Networks (CNNs) \cite{PixelWise, laserpowder, RandomRef}.

% \IEEEPARstart{S}{urface}
%[2, 3, 4].

\begin{figure}[htbp]
\centerline{\includegraphics[width=8cm]{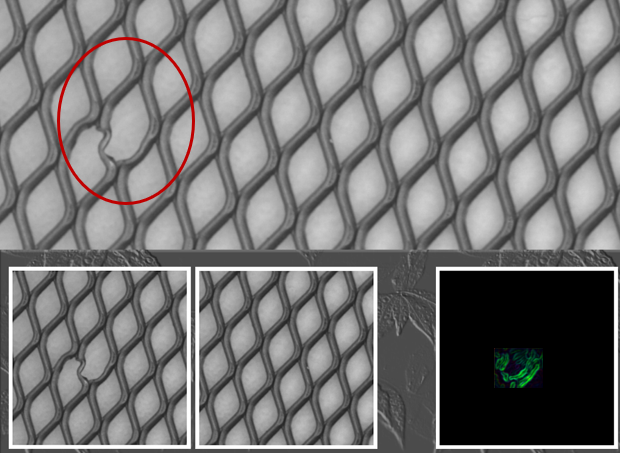}}
\caption{Test case surface example. The three images below show the defective patch, the prototype patch and the resulting pixel-wise anomaly score as learnt by the CNN.}
\label{fig:IntroFigure}
\end{figure}

A CNN is a class of deep neural networks that uses convolution in place of general matrix multiplication in at least one of their layers, and reduces the number of parameters very efficiently without losing out on the quality of models, making it a prime choice for analyzing visual imagery. A basic requirement for training CNNs efficiently on a classification objective is the accessibility of an adequately large amount of training data for each class. However, with well-optimized processes, there is often an abundance of non-defective samples and a relatively small amount of defective samples. To address this data imbalance challenge, the training objective can be shifted from defect classification to anomaly detection, eliminating the need for any defective samples for training \cite{ADSurvey}. An additional advantage of this approach is potentially detecting anomalies in novel classes that are not part of the training data set, providing a general solution to the surface quality inspection task.

Some previous approaches in the literature use defective samples for training, or involve training the CNN features to assign a lower feature-space distance to similar parts and a higher feature-space distance to dissimilar parts based on solving a classification task, not solving the anomaly detection task described in this work. Other approaches involve detecting anomalies in certain regions of an image, not capturing potential defects across the entire image.

% focuses on detecting anomalies in certain regions of an image, not capturing potential defects across the entire image. Other approaches involve training the CNN features to assign a lower feature-space distance to similar parts and a higher feature-space distance to dissimilar parts based on solving a classification task, rather than directly learning a similarity metric for distance-based surface anomaly detection.

In this paper, efficient CNNs based on residual networks \cite{ResNet} are trained to explicitly learn a similarity metric for defective and non-defective patches of surface textures through the use of a triplet network model \cite{TripletNetwork}. The defective samples used in the triplet network are artificially augmented exclusively from non-defective samples via random erasing \cite{RandomDataErasing} techniques. After training, a simple prototype pixel-wise subtraction method is performed between the evaluated surface and the prototype measurement for that particular texture surface in the feature space learned by the CNN, producing an array of distance values (see Fig. \ref{fig:IntroFigure}). A Support Vector Machine (SVM) with a simple linear kernel is then applied against the mean and maximum distance values collected, with the intention to maximize the separation between the defective and non-defective evaluated samples. Anomaly detection for known classes that are part of the training data set, as well as novel classes that are not part of the training data set, are explored for the industrial surfaces. In addition to the surfaces, the different defect types present in the testing batch are explored individually, to gain a deeper insight into the anomalies perceptibility.

% \begin{figure}[htbp]
% \centerline{\includegraphics[width=8cm]{"figures/Photoshop with Heatmap 2".png}}
% \caption{Test case surface example. The three images below show the defective patch, the ideal prototype patch and the resulting pixel-wise anomaly score as learnt by the CNN.}
% \label{fig:IntroFigure}
% \end{figure}

%to be evaluated against a particular threshold. 

The remainder of this paper is structured as follows. Section \ref{relatedwork} presents the motivations behind the use of machine learning and the related work in the field of industrial surface anomaly detection. Section \ref{dataset} contains background information about the data set used. Section \ref{methodology} details the methodology and implementation. Section \ref{results} discusses the obtained results and performance evaluation. Finally, Section \ref{conclusion} concludes the paper and discusses opportunities for future work.

\section{Motivation and Related Work}\label{relatedwork}
The following outline some of the work being done in the field related to the use of machine learning in anomaly detection of industrial surface textures. 

Haselman, \emph{et al.} \cite{ADusingDL} propose a one-class unsupervised learning on fault-free samples by training a deep CNN to complete images whose center regions are cut out. Results demonstrate the approach's suitability for detecting visible anomalies, but fails to accurately detect weakly contrasted anomalies and focuses on the center regions of an image only. Chai, \emph{et al.} \cite{ExploitingSparsity} propose an approach that utilizes sparse representation to identify anomalies obtained during image reconstruction. Since sparse representation is a computationally expensive approach, lower image resolutions are used, which in turn lowers the accuracy of the anomaly detection task. Natarajan, \emph{et al.} \cite{CNNVotingBased} propose the use of transfer learning and a majority voting mechanism for image representations and fuse extracted features. This approach still requires defective samples during training to achieve the anomaly detection objective. Napoletano, \emph{et al.} \cite{ADNanofibrous} propose a region-based method for detecting and localizing anomalies via evaluating the degree of abnormality of each subregion of an image compared to an anomaly-free image. The CNN features are trained to specifically assign a low distance to similar parts and a high distance to dissimilar parts based on a classification task, rather than an anomaly detection objective.

The previous work of Staar, \emph{et al.} \cite{ADStaar} tackles the shortcomings of the work mentioned above, by using Gaussian noise to synthesize defective samples from non-defective images and utilizing a residual-based \cite{ResNet} triplet network \cite{TripletNetwork}. The network directly learns a similarity metric for the distance-based surface anomaly detection objective, where same-class samples have lower feature space distances than out-of-class samples. Defective images are separated from non-defective images during testing via varying threshold values against the maximum distance value collected. 

The work presented in this paper further extends Staar, \emph{et al.} \cite{ADStaar} methodology and results, while training and testing on a real-world data set with a focus on industrial surfaces rather than an artificially generated general texture data set. 

The main contributions of this paper include:
\begin{itemize}
\item Artificially synthesize defective images exclusively from non-defective samples using lightweight random erasing techniques that do not require any extra memory consumption or parameter learning. 
%and the use of defective images from the data set is eliminated.
% \item Defective images are artificially synthesized using random data erasing \cite{RandomDataErasing} rather than Gaussian noise. Pixels of the erased region are not re-assigned with any noise.
%as it translates well to the deviations in the data set used..... re-assigned with random noise
\item Implement a deep, residual-based CNN architecture with very small convolution kernels. This allows the network to learn more interesting features while having a small number of trainable parameters.
% \item A deeper, more efficient CNN architecture with very small convolution kernels is implemented, allowing the network to learn more interesting features.
%while reducing the number of trainable parameters.
%which is inspired by some of the state-of-art network architectures available today.
\item Maximize the margin of separation between the defective and non-defective images during evaluation, by utilizing a hard-margin SVM against both the mean and maximum feature space distance values collected.
% \item SVM is introduced during evaluation against the mean and maximum distance values collected, to maximize the separation between the defective and non-defective images. as opposed to varying threshold values against the maximum distance values only.
\item Analyze the individual defect types to gain a deeper insight into the anomalies' perceptibility when using the proposed approach.
% \item Defect types are analyzed individually to gain a deeper insight into the anomalies' perceptibility when using this paper’s approach.
\end{itemize}

% Training utilizes just 64x64 patches of an image rather than the entire image, where the patches are of different orientations and locations.

% Our approach differs from the works presented above as no defective images are used in training, it does not rely on transfer learning, it utilizes just patches of an image rather than the entire image, it can be generalized for novel detection, and CNN features are not trained to specifically assign low distance to similar parts and high distance to dissimilar parts.

% The optimization techniques include exploring different loss functions, distance measures, network hyperparameters, data pre-processing techniques, and a network architecture, as well as introducing SVM to maximize the separation between the defective images and non-defective images during evaluation, as opposed to varying threshold values.

\section{Background}\label{dataset}

\begin{figure}[htbp]
\centerline{\includegraphics[width=8cm]{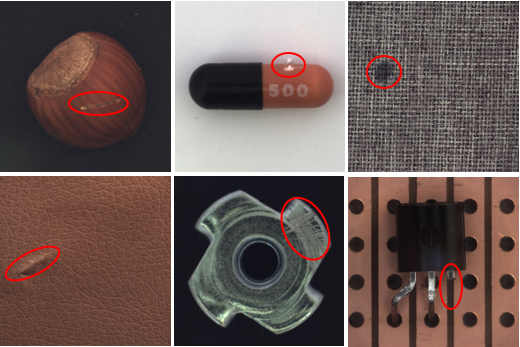}}
\caption{Defective samples from the MVTec AD \cite{mvtec}. The defective positions are marked in red.}
\label{fig:MVTec}
\end{figure}

The data set used in this paper is the MVTec Anomaly Detection Dataset (MVTec AD) \cite{mvtec}. It is a modern, comprehensive, real-world data set with over 5000 images, divided into 15 different industrial object and texture category classes. Each class consists of non-defective data and a range of different types of defective data, such as bent, broken, and contaminated surfaces. Image resolutions lie between 700x700 and 1024x1024 pixels. In order to evaluate anomaly detection for both known and novel classes, 11 classes are used in training and testing, whereas the remaining 4 classes are only used in testing. The classes are pre-selected randomly. Of the classes to be trained, half of their non-defective images are used for training, and the other half are available for testing. Figure \ref{fig:MVTec} shows a sample of defective images from the data set and marks the defective positions in red. 

% Each class consists of non-defective data as well as different types of defective data such as bent, broken, and contaminated surfaces.

% All the defective images, except 'color' defect types, are used for testing. 'Color' defects are omitted as all images are converted to grayscale during data pre-processing, which would mask any RGB defects. 

% The second data set is the Kylberg Texture Dataset v1.0 \cite{kylberg}. It consists of 28 texture classes, such as fabrics, surfaces of stone, rice grains, and sesame seeds. Each class contains 160 grayscale images at 576x576 pixels resolution. All images will be used for training.

% (Will I still use Kylberg? Explore other texture/object data sets)

\section{Methodology}\label{methodology}

The following section details the model's design, implementation and evaluation metrics.

\subsection{Model}

The triplet network model is utilized in this paper for training \cite{TripletNetwork}. This architecture allows an input of three different images to be fed into the same CNN, where the weights are shared. The first image is called the anchor (\emph{a}) image, the second image is called the positive (\emph{p}) image, and the third image is called the negative (\emph{n}) image. The \emph{a} and \emph{p} images belong to the same class, and the \emph{n} image belong to a different class. Features of each image are computed, before calculating the Euclidean distance (\emph{d1}) between the resulted features of \emph{a} and \emph{p} and the Euclidean distance (\emph{d2}) between the resulted features of \emph{a} and \emph{n}. This allows the network to directly learn a similarity metric via deep metric learning, which is beneficial as the number of object classes is not specified and could be boundless. The Euclidean distance is used as initial experiments show better results when it is employed compared to the Manhattan and the Minkowski distance metrics. Equations \ref{equ:eucpos} and \ref{equ:eucneg} show the Euclidean distance measurement for \emph{d1} and \emph{d2} respectively.

\begin{equation}
    d1 = \sqrt {\sum\limits_{i=1}^n (a_i-p_i)^2}
    \label{equ:eucpos}
\end{equation}

\begin{equation}
    d2 = \sqrt {\sum\limits_{i=1}^n (a_i-n_i)^2}
    \label{equ:eucneg}
\end{equation}

\begin{figure}[htbp]
\centerline{\includegraphics[width=8cm]{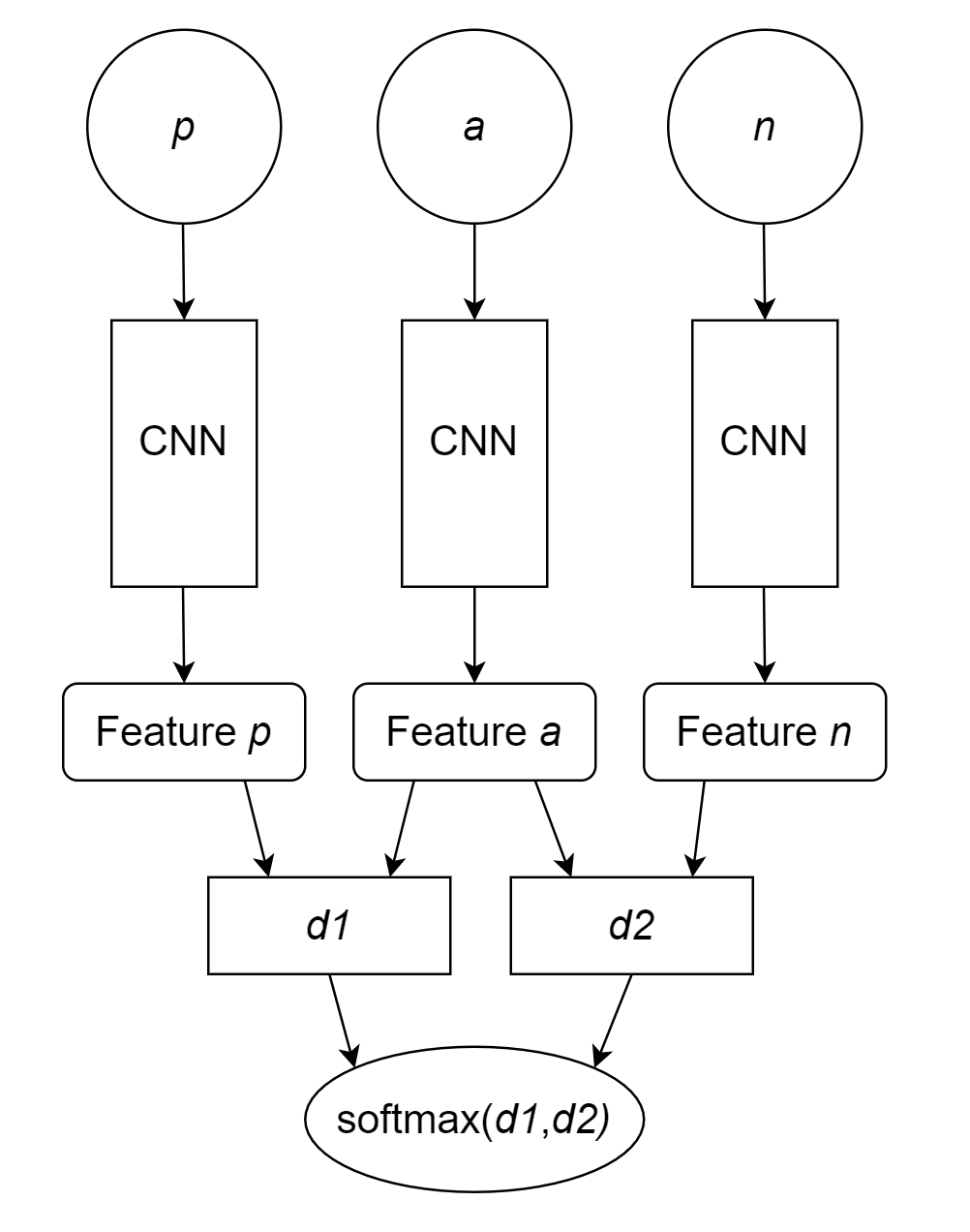}}
\caption{Training Model}
\label{fig:TripletNetwork}
\end{figure}

Afterwards, the resulting \emph{d1} and \emph{d2} scores are inserted into a row vector then placed into the softmax function, yielding an output vector who's sum is equal to one. The output vector is trained with the target vector [1,0] for each triplet of (\emph{a}, \emph{p}, \emph{n}), ensuring the network assigns a higher similarity score between images \emph{a} and \emph{p} and a lower similarity score between images \emph{a} and \emph{n}. Figure \ref{fig:TripletNetwork} visualizes the complete training model.

\subsection{Data Pre-Processing}

Before training the network, all training images are resized to 1024x1024 pixel resolutions, converted to grayscale, and normalized via Min-Max scaling to rescale their values between 0 and 1. Feature scaling eliminates a large scaled feature to be dominating, as the Euclidean distance measure is sensitive to magnitudes, while preserving all relationships in the data \cite{featurescaling}. Furthermore, it allows gradient descent to converge much faster.

% all images from the Kylberg data set are resized to 512x512 pixel resolution and

% \begin{figure}[htbp]
% \centerline{\includegraphics[width=9cm]{"figures/Image Resizing".png}}
% \caption{Left half: Original image. Right half: 64x64 patches extracted from various input sizes of the original image.}
% \label{fig:ImageResizing}
% \end{figure}

The CNN implemented in this paper is only required to learn similarities between patches of 64x64 pixels. The idea is inspired by previous work of Weimer, \emph{et al.} \cite{automatedFE} where 32x32 pixels is sufficient for defect classification of images of 512x512 pixel resolution. As a generalization, 64x64 pixels would be sufficient for images of 1024x1024 pixel resolution. In addition, the use of small image patches extracted from an image enables the amount of training data to be increased. To keep the patch size of 64x64 fixed and to capture textural regularities that are expressed on different spatial scales, the input training images are randomly resized to either 1024x1024, 512x512, 256x256 or 128x128 pixel resolutions. 16 random patches are then extracted from the 1024x1024 and 512x512 pixel resolution images, and 8 random patches are extracted from the 256x256 and 128x128 pixel resolution images. 

%Figure \ref{fig:ImageResizing} visualizes a sample image and various 64x64 patches extracted from its different resized values.

% \begin{figure}[htbp]
% \centerline{\includegraphics[width=9cm]{"figures/Image Resizing July".png}}
% \caption{Left half: Original image. Right half: 64x64 patches extracted from various input sizes of the original image.}
% \label{fig:ImageResizing}
% \end{figure}

%  Kylberg input training images are randomly resized to either 512x512, 256x256 or 128x128 pixel resolutions, 

\begin{figure}[htbp]
\centerline{\includegraphics[width=9cm]{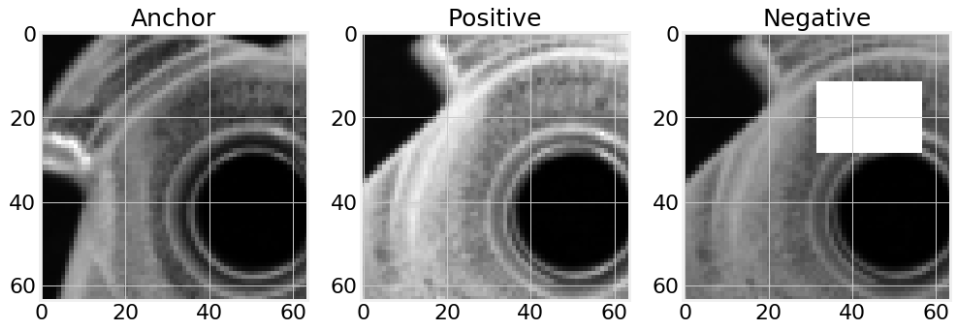}}
\caption{Triplet input sample of (\emph{a}, \emph{p}, \emph{n})}
\label{fig:TripletInputSample}
\end{figure}

The network is trained without any defective images on an anomaly detection objective, due to the limited number of defective samples to train the network efficiently on a classification objective and to further generalize the solution to the surface inspection task. Defective images used as the \emph{n} image input into the network are instead artificially synthesized exclusively from non-defective images using random erasing \cite{RandomDataErasing}. The random erasing technique utilized essentially erases a random part of an image of a random size, enhancing the sensibility to local deviations. Random erasing has the advantage of being a lightweight method, as it does not require any extra memory consumption or parameter learning. Figure \ref{fig:TripletInputSample} visualizes an input sample of (\emph{a}, \emph{p}, \emph{n}) to the triplet network.

%surface quality inspection task%
% It is applied to same-class samples, and the synthesized image is inserted into the network as the \emph{n} image.

\subsection{Network Architecture}
The CNN implemented in this paper is based on deep residual learning \cite{ResNet} and uses very small convolution kernels, inspired by the work of the widely-known VGG16 network \cite{VGG16}. Residual networks ease the training and optimization of networks, by enabling layers to learn a residual function referenced to the input layer as opposed to learning unreferenced functions. Moreover, the skip connections between layers add the outputs from previous layers to the outputs of stacked layers, allowing networks to become deeper. Deeper networks allow more interesting hierarchical features to be learned and can produce a better generalization \cite{DeeperNetworksRELU}. In addition, the VGG16 network \cite{VGG16} demonstrates how a deeper network with very small (3x3) convolution kernels produce more accurate results compared to the previous configurations on the ImageNet Large Scale Visual Recognition Challenge 2014 (ILSCVRC2014) \cite{ILSVRC2014} that use larger convolution kernels. Small convolution kernels significantly reduce the number of parameters in the network, where a single 3x3 convolution kernel has a number of parameters ratio of 1:1.4 with a 5x5 convolution kernel, and a 1:2 with a 7x7 convolution kernel.

%  Deeper networks allow more interesting features to be learned, as opposed to wider networks that have too many weights and may overfit.

\begin{figure}[htbp]
\centerline{\includegraphics[width=9cm]{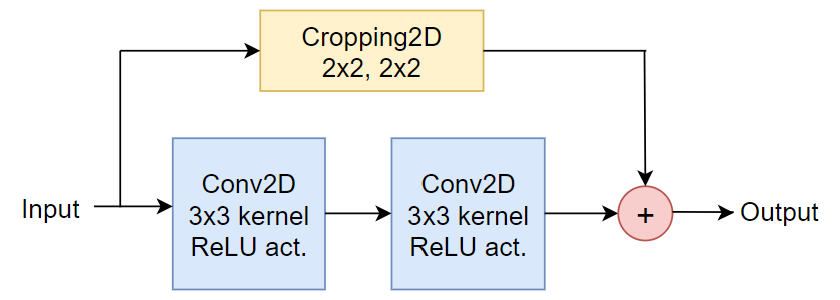}}
\caption{A residual block used in the network.}
\label{fig:ResidualBlock}
\end{figure}

The first 3 layers in the network presented in this work consist of 2D convolutional layers with Rectified Linear Unit (ReLU) activation functions and kernel sizes of 3x3, followed by a 2D maxpool layer with a window size of 2x2 and strides 2x2. Afterwards, seven residual blocks are stacked next to each other. Each block consists of two consecutive 2D convolutional layers with ReLU activation functions and kernel sizes of 3x3 applied on the input, which produce an output spatial resolution that is reduced by 4x4. Their output is then added to the output of a 2D cropping layer, which applies a spatial dimension crop of size 2x2 to the height and 2x2 to the width of the input. That is required so the output spatial dimension matches that of the 2D convolutional layers, in order for the learned residual to be added and produce the final residual block output. Figure \ref{fig:ResidualBlock} visualizes a residual block setup and parameters used in the network.

% A residual block is the basic unit of a residual network. The residual block presented in this work consists of two consecutive 2D convolutional layers with Rectified Linear Unit (ReLU) activation functions and kernel sizes of 3x3 applied on the input, which produce an output spatial resolution that is reduced by 4x4. Their output is then added to the output of a 2D cropping layer, which applies a spatial dimension crop of size 2x2 to the height and 2x2 to the width of the input. That is required so the output spatial dimension matches that of the 2D convolutional layers, in order for the learned residual to be added and produce the final residual block output. ReLU is utilized as it solves the vanishing gradient problem and is very efficient to compute, as it is linear for positive values and zero for negative values \cite{DeeperNetworksRELU}. Figure \ref{fig:ResidualBlock} visualizes a residual block setup and parameters used in the network.
 
%  The network presented in this work comprises of residual blocks. Each block

%The input goes through two consecutive 3x3 2D convolution kernels with ReLU activations, and goes through a 2D cropping layer of 2x2 applied to the height and 2x2 applied to the width. Their outputs are then added to each other, producing the final residual block output.

Higher-level features are learnt with every layer and residual block. Besides the 3x3 kernel, all convolutional layers across the network use 16 number of filters, valid padding option, and the ReLU activation function. ReLU is utilized as it solves the vanishing gradient problem and is very efficient to compute, as it is linear for positive values and zero for negative values \cite{DeeperNetworksRELU}. The entire network is built fully convolutional to allow weights to be shared between the training and testing models, as they process different image pixel resolutions. During training, 64x64 pixels are transformed to tensors with dimensions of 1x1x16, whereas during testing, 1024x1024 pixels are transformed to tensors with dimensions of 481x481x16.

\subsection{Network Implementation and Hyperparameters}

All networks are built and implemented in Python 3.7.4, using the Tensorflow \cite{Tensorflow} and Keras \cite{Keras} libraries.

For training, the Adam optimizer is utilized with a fixed learning rate of 0.0001 and a batch size of 256 samples. The mean absolute error (MAE) loss function is utilized, as initial experiments show much better results with it compared to the mean squared error (MSE) and the categorical cross entropy loss functions. The network is trained for 40 epochs. 

% x 4 times, where a different set of triplets are generated each time to allow the network learn different input variations.

\subsection{Evaluation Metrics}

As for defect detection evaluation, a class prototype is first created by embedding a large amount of non-defective data of that class into the CNN to obtain a set of features. Afterwards, the testing images are embedded into the same CNN and the Euclidean distances between the prototype and the testing images features are calculated. The resulted maximum and mean distances are then subjected to an SVM with a linear kernel, with the intention of maximizing the separation between the defective test images and the non-defective test images. The maximum and mean Euclidean distances were selected based on recent work which addressed the setting of working-point decision thresholds for an anomaly detection task \cite{Aburakhia}. An SVM with a linear kernel allows the robust classification algorithm to be fast and memory efficient, as it is a single inner product with a O(\emph{d}) prediction complexity, where \emph{d} is the number of input dimensions. The regularization parameter (\emph{C}) is kept constant at 1E8, for a harder-margin SVM that imposes a larger penalty for wrongly classified examples.

% The linear kernel works well as the data points are close to being linearly separable.

\begin{figure*}[htbp]
\centerline{\includegraphics[width=18cm]{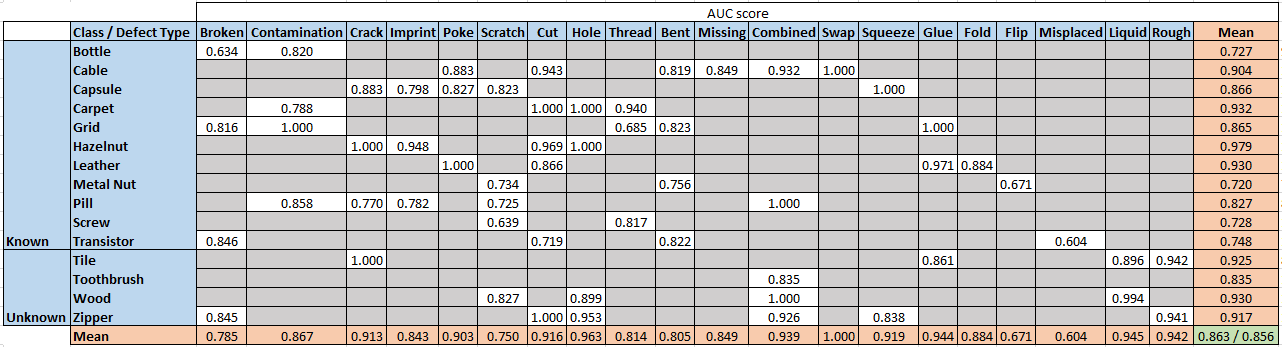}}
\caption{Staar, \emph{et al.} \cite{ADStaar} approach AUC scores of discrimination between the defective and non-defective images.}
\label{fig:Staar Results Table}
\end{figure*}

\begin{figure*}[htbp]
\centerline{\includegraphics[width=18cm]{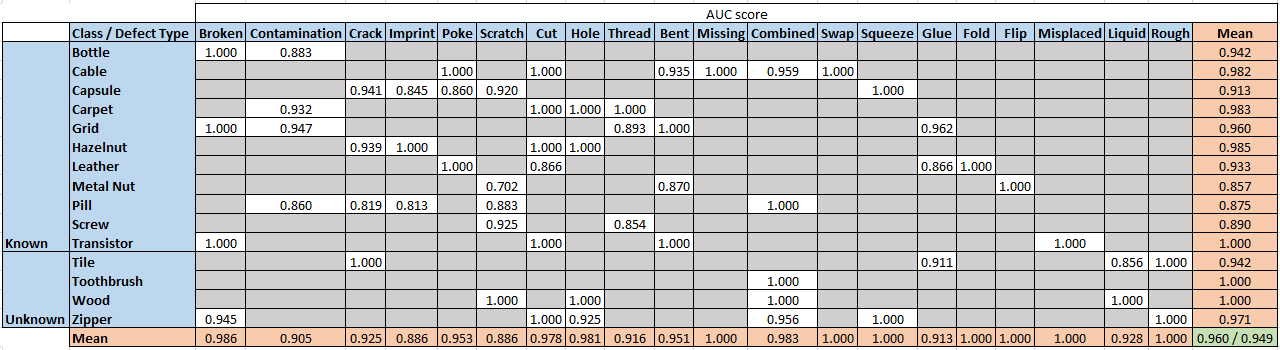}}
\caption{The proposed approach AUC scores of discrimination between the defective and non-defective images.}
\label{fig:Results Table}
\end{figure*}

%for all classes and defect types.

To evaluate how well the proposed approach discriminates between the defective and non-defective images for each surface and the individual defect types, the Area Under the Curve (AUC) of the Receiver-Operating Characteristic (ROC) is calculated. The higher the AUC, the better the model is at predicting true positives and true negatives. True positives in this work indicate defective images correctly classified as defective and true negatives indicate non-defective images correctly classified as non-defective. Experiments are repeated three times and the average AUC score is recorded. As the work presented in this paper extends the approach used by Staar, \emph{et al.} \cite{ADStaar}, their approach is tested against the MVTec AD \cite{mvtec} for a comparison of results.

% Staar, \emph{et al.} \cite{ADStaar} approach is tested against the MVTec AD \cite{mvtec} for a comparison of results.

% \begin{figure*}[htbp]
% \centerline{\includegraphics[width=18cm]{"figures/Staar Results with average".png}}
% \caption{Staar, \emph{et al.} \cite{ADStaar} AUC scores of discrimination between the defective and non-defective images for all classes and defect types.}
% \label{fig:Staar Results Table}
% \end{figure*}

% \begin{figure*}[htbp]
% \centerline{\includegraphics[width=18cm]{"figures/Our Results with average".png}}
% \caption{The paper's AUC scores of discrimination between the defective and non-defective images for all classes and defect types.}
% \label{fig:Results Table}
% \end{figure*}

\section{Performance Evaluation}\label{results}

Figures \ref{fig:Staar Results Table} and \ref{fig:Results Table} show the individual and mean AUC scores for Staar, \emph{et al.} \cite{ADStaar} approach and the proposed approach, respectively. Upon examining the table of results, the proposed approach produces a better AUC score for the majority of the individual class-defect types, and a better mean AUC score for every single class and defect type. The proposed approach scores an average class mean AUC of 0.093 higher, and an average defect type mean AUC of 0.097 higher.

Furthermore, it can be observed from the class mean scores that the proposed approach achieves some state-of-art results for both known classes and novel classes. In total, only three classes achieved a mean AUC score less than 0.90, four classes achieved a mean AUC score between 0.90 and 0.95, and the remaining eight classes achieved a mean AUC score between 0.95 and 1.0. In particular, three classes achieved a perfect mean AUC score of 1.0, where only one class was known and two were novel, further demonstrating how well the method translates to unknown inputs.

Regarding defect types, it can be seen from their mean scores that the proposed approach achieves some state-of-art results. Out of the twenty different defect types, only imprint and scratch defect types achieved a mean AUC score less than 0.90. Contamination, crack, and thread defect types achieved a mean AUC score between 0.90 and 0.95, and the remaining fifteen defect types achieved a mean AUC score between 0.95 and 1.0.

\begin{figure}[htbp]
\centerline{\includegraphics[width=8.8cm]{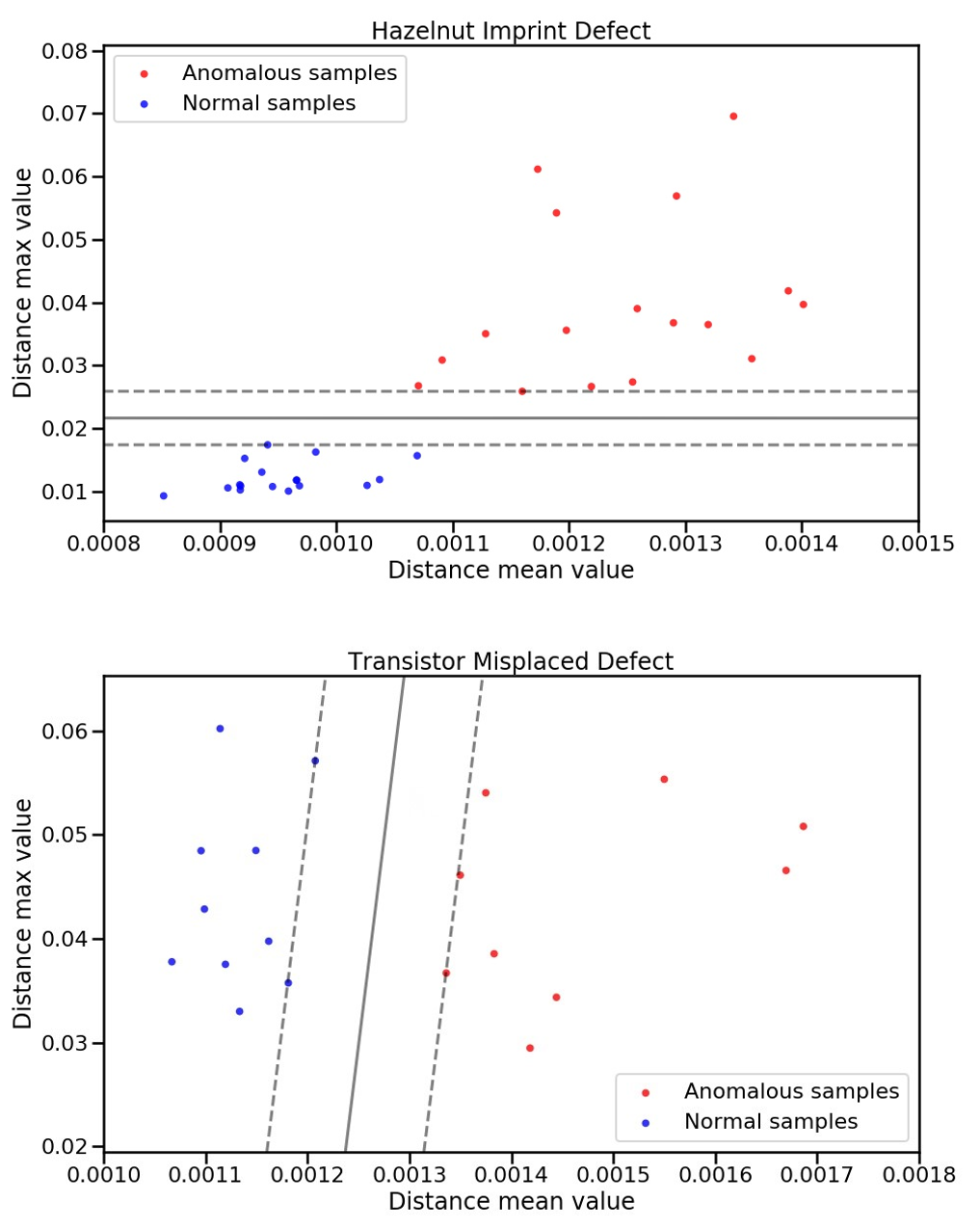}}
\caption{Scatter plot examples. Top: Hazelnut class with imprint defects. Bottom: Transistor class with misplaced defects.}
\label{fig:ScatterPlot}
\end{figure}

Figure \ref{fig:ScatterPlot} shows two scatter plot examples with perfect linear SVM separation between the defective and non-defective samples. The plot at the top shows a separation mainly via features max distance with the prototype, and the plot at the bottom shows a separation mainly via features mean distance with the prototype. 

% \begin{figure}[htbp]
% \centerline{\includegraphics[width=10cm]{"figures/hazelnut_imprint_defect".jpg}}
% \caption{Scatter plot examples. Top: Hazelnut class with imprint defects. Bottom: Leather class with fold defects.}
% \label{fig:ScatterPlot}
% \end{figure}

% (AUC-ROC curve)

% (Maybe include images of heat map / localization)

\section{Conclusion}\label{conclusion}

This paper demonstrated how deep metric learning can be effectively used for surface anomaly detection of different defect types without the use of any defective samples. It further showed how this approach translated well to novel classes that were not part of the training data set, as two of the four novel classes achieved a perfect AUC score of discrimination between the defective and non-defective images. 

As for the methodology, a triplet network model was utilized, which included a deep residual-based CNN with very small convolution kernels. A modern, comprehensive, real-world data set with a focus on industrial surfaces was used for training and testing, where training defective samples were synthesized exclusively from the non-defective samples via random erasing. Image patches of 64x64 were utilized for training instead of using the entire image, allowing training to be more efficient.

Results were promising, but they could still be improved for certain classes and defect types. A potential limitation could be for object classes, where an object comprises a small area of the entire image and most of the extracted 64x64 patches are of the background rather than the object itself. That is backed up by the fact that pill and screw object classes were two of the three classes that achieved a mean AUC score of less than 0.90, where they had the smallest object to whole image ratio in the entire data set. Furthermore, it can be seen that some classes had more defective images than others, and some defect types were present in more classes than others, resulting in an unbalanced comparison of results between the different classes and defect types.

Future work include exploring the addition of training data from other texture and object data sets, as that can allow the network to learn more powerful and complex features. In addition, more testing data could be obtained to gain a deeper understanding and more accurate defect types evaluations. The results could be further enhanced through the use of batch normalization in the CNN, which reduces internal covariate shifts and can accelerate deep network training. The use of maxpooling layers and convolutional layers with same padding could be explored instead of the use of convolutional layers with valid padding. Other potential enhancements include exploring additional loss functions and distance metrics, larger patch sizes, smarter extraction of object patches for surface images with a small object area coverage, different data augmentation techniques, and further tuning the network hyperparameters \cite{LiShami}.

% The CNN could be further enhanced through the use of batch normalization \cite{Batchnorm}, which reduces internal covariate shifts and can accelerate deep network training.

% (Include References if possible later on)

% (This could be attributed to the fact that some of those defects are even hard for a human being to locate / these are the hardest for the CNN to mathematically locate or compute due to some samples being / Give a mathematical reasoning if possible for high/low AUC scores for both classes and defect types)

% or


% \appendices
% \section{Group Members Contribution}
% Each member contributed equally to all parts of the project. This includes the literature research and survey, data set exploration, data pre-processing, models building and implementations, and result comparisons. Meetings were conducted every week via zoom to discuss our project progress. 

% \section{}

\section*{Acknowledgment}
This work is partially supported by the Natural Research Council (NRC) of the Government of Canada under Project AM-105-1.

\ifCLASSOPTIONcaptionsoff
  \newpage
\fi

\end{document}